%% file: Neighbortrack_arxiv.tex
\documentclass[10pt,onecolumn,letterpaper]{article}
\usepackage{graphicx}
\graphicspath{ {./figures/} }
\usepackage{amsmath}
\usepackage{amssymb}
\usepackage{booktabs}
\usepackage[section]{placeins}
\usepackage[T1]{fontenc}    % use 8-bit T1 fonts
\usepackage[utf8]{inputenc} % allow utf-8 input
\usepackage{arxiv}
\usepackage{url}            % simple URL typesetting
\usepackage{amsfonts}       % blackboard math symbols
\usepackage{nicefrac}       % compact symbols for 1/2, etc.
\usepackage{microtype}      % microtypography
\usepackage{lipsum}
\usepackage{graphicx}
\graphicspath{ {./figures/} }
\usepackage{CJKutf8}
\usepackage{verbatim}
\usepackage[ruled]{algorithm2e}                 %演算法排版樣式1
\usepackage{color}
\usepackage{ulem}
\usepackage{xcolor}
\usepackage{bm}
\usepackage{authblk}

\usepackage[pagebackref,breaklinks,colorlinks]{hyperref}

\usepackage[capitalize]{cleveref}
\usepackage{arxiv}

\crefname{section}{Sec.}{Secs.}
\Crefname{section}{Section}{Sections}
\Crefname{table}{Table}{Tables}
\crefname{table}{Tab.}{Tabs.}

\begin{document}

\title{NeighborTrack: Improving Single Object Tracking by Bipartite Matching with Neighbor Tracklets}
\author[1]{Yu-Hsi Chen}
\author[1]{Chien-Yao Wang}
\author[2]{Cheng-Yun Yang}
\author[1]{Hung-Shuo Chang}
\author[3]{Youn-Long Lin}
\author[4]{Yung-Yu Chuang}
\author[1]{Hong-Yuan Mark Liao}
\affil[1]{Institute of Information Science, Academia Sinica, Taiwan, \textit {\{franktpmvu,kinyiu,jonathanc,liao\}@iis.sinica.edu.tw}}
\affil[2]{Department of Electrical Engineering, Purdue University, \textit {yang2316@purdue.edu}}
\affil[3]{Department of Computer Science, National Tsing Hua University, \textit {ylin@cs.nthu.edu.tw}}
\affil[4]{Department of Computer Science and Information Engineering, National Taiwan University, \textit {lcyy@csie.ntu.edu.tw}}

\maketitle

\input{macros}

\input{0abstract}
\input{1intro}
\input{2related}

\input{3method}
\input{4experiments}

\input{5conclusion}
\input{6additional_result}

%%%%%%%%% REFERENCES
{\small
\bibliographystyle{ieee_fullname}
\bibliography{Neighbortrack_arxiv}
}

\end{document}

%% file: macros.tex
\definecolor{green}{rgb}{0, 0.6, 0} 

\newcommand{\heading}[1]{\noindent\textbf{#1}}

\newcommand{\figref}[1]{Figure~\ref{fig:#1}}% type "\figref{}" to reference figure
\newcommand{\tabref}[1]{Table~\ref{tab:#1}} % type "\tabref{}" to reference table
\newcommand{\itmref}[1]{[\ref{itm:#1}]}     % type "\itmref{}" to reference item
\newcommand{\eqnref}[1]{Equation~\ref{eq:#1}}
\newcommand{\secref}[1]{Section~\ref{sec:#1}}

%% editing comment
\newcommand{\ignore}[1]{}   % ignore this
\newcommand{\cmt}[1]{\begin{sloppypar}\large\textcolor{red}{#1}\end{sloppypar}}

\newcommand{\TODO}[1]{\textcolor{red}{[TODO]\{#1\}}}
\newcommand{\todo}[1]{\textcolor{red}{#1}}
\newcommand{\torevise}[1]{\textcolor{blue}{#1}}
\newcommand{\revise}[1]{\textcolor{blue}{#1}}
\newcommand{\copied}[1]{ \textcolor{red}{[COPIED: #1]}}
\newcommand{\cyy}[1]{\textcolor{green}{#1}}

%% file: 0abstract.tex
\begin{abstract}

We propose a post-processor, called NeighborTrack, that leverages neighbor information of the tracking target to validate and improve single-object tracking (SOT) results. 
It requires no additional data or retraining.
Instead, it uses the confidence score predicted by the backbone SOT network to automatically derive neighbor information and then uses this information to improve the tracking results.
When tracking an occluded target, its appearance features are untrustworthy.
However, a general siamese network often cannot tell whether the tracked object is occluded by reading the confidence score alone, because it could be misled by neighbors with high confidence scores.
Our proposed NeighborTrack takes advantage of unoccluded neighbors' information to reconfirm the tracking target and reduces false tracking when the target is occluded.
\iffalse
It not only reduces the impact caused by occlusion, but also fixes tracking problems caused by object appearance changes.
NeighborTrack is agnostic to SOT networks and post-processing methods. 
\fi
For the VOT challenge dataset commonly used in short-term object tracking, we improve three famous SOT networks, Ocean, TransT, and OSTrack, by an average of ${1.92\%}$ EAO and ${2.11\%}$ robustness.
For the mid- and long-term tracking experiments based on OSTrack, we achieve state-of-the-art 72.25${\%}$ AUC on LaSOT and 75.7${\%}$ AO on GOT-10K. Code duplication can be found in https://github.com/franktpmvu/NeighborTrack.

%We propose a post-processor, called NeighborTrack, that leverages neighbor information of the tracking target to validate and improve single-object tracking (SOT) results.  
%It requires no additional verification data nor retraining.
%Instead, it uses the confidence score output from the baseline network to automatically derive neighbor information, and then use this information to improve the tracking results.
%When tracking an occluded target, its appearance features are untrustworthy.
%However, a general siamese network \cite{siameseFCbertinetto2016fullyconvolutional} cannot tell whether the tracked object is occluded by reading the confidence score alone, because it could be misled by neighbors with high confidence score.
%Our proposed NeighborTrack takes advantage of unoccluded neighbors' information to reconfirm the tracking target and reduce false tracking when the target is occluded.
%It not only reduces the impact caused by occlusion, but also fixes tracking problems caused by object appearance changes.
%We verify NeighborTrack using multiple baseline networks and post-processing methods. 
%For the VOT challenge dataset\cite{VOT_TPAMI} commonly used in short-term object tracking, we improve three famous networks, Ocean\cite{ocean_zhang2020}, TransT\cite{TransT}, and OSTrack\cite{ostrack}, by an average of ${1.92\%}$ EAO and ${2.11\%}$ robustness.
%For the mid- and long-term tracking experiments based on OSTrack\cite{ostrack}, we achieve state-of-the-art 72.25${\%}$ AUC on LaSOT\cite{lasotDBLP:journals/corr/abs-1809-07845} and 75.7${\%}$ AO on GOT-10Ks.

\end{abstract}

%% file: 1intro.tex
\section{Introduction}
\label{sec:intro}

Single Object Tracking (SOT) is a fundamental computer vision task that establishes the correspondence of an arbitrarily specified object along time~\cite{siammask_wang2019fast}.
There are numerous applications for it, including video surveillance~\cite{visual_survalance_Mangawati_2018,visual_survalance_Xing_2010}, video annotation~\cite{Semi_Automatic_Annotation_Berg_2019_ICCV}, human-computer interaction~\cite{handtrack_liu_icpr2010}, etc.
An SOT network takes a user-specified target of interest in the first frame and then tracks its position in subsequent frames.
In contrast to multi-object tracking (MOT), which knows the target classes in advance, SOT is unaware of the target classes.
The current SOT approaches use various algorithms derived from the Siamese network~\cite {siameseFCbertinetto2016fullyconvolutional}.
With the help of a deep learning network, they extract the appearance feature of the target object and then use that feature to locate the positions of objects with similar features in subsequent frames.
Using pairwise appearance feature alignment, the same network can be used to track multiple objects with different appearance features without requiring retraining.
This approach, however, might fail to track when the appearance characteristics of either side change.

In recent years, deep neural network-based feature extraction has led to an increase in the accuracy of tracking objects.
It is, however, unreliable to track objects purely based on their appearance features, as their appearance can change due to deformations, changes in size, changes in color, and occlusions.
In particular, occlusion presents a serious challenge since we cannot predict what objects are occluding the target or are occluded by the target, and therefore cannot use the partial appearance to continue tracking.
Further, when tracking one among several similar objects, occlusion may cause the tracker to assume an incorrect position. 

\begin{figure*}[t]
  \centering
  %\fbox{\rule[-.5cm]{4cm}{4cm} \rule[-.5cm]{4cm}{0cm}}
  \includegraphics[width=0.7\linewidth]{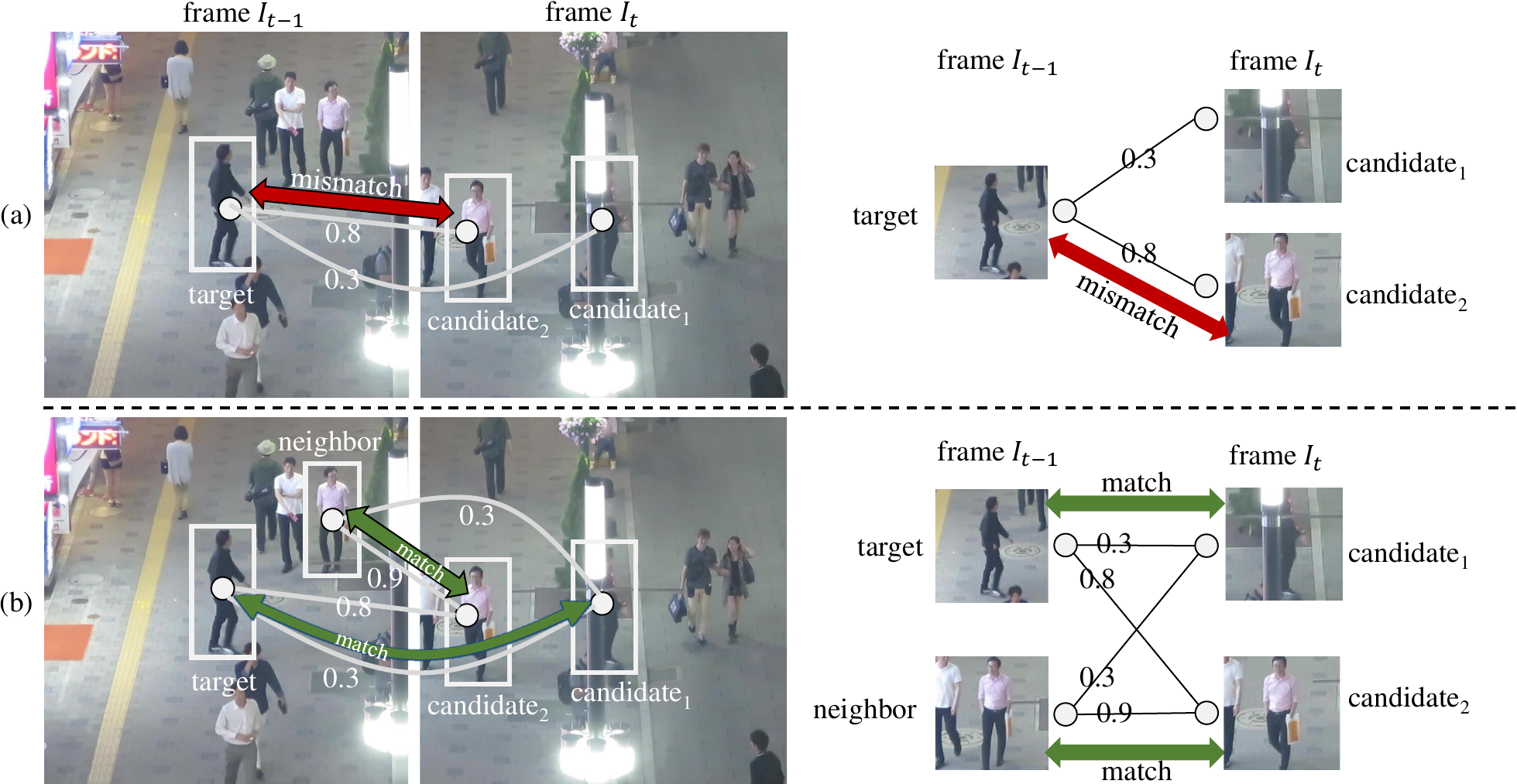}
  \caption{An illustrative example of how neighbor information can be used to correct tracking results. 
  Neighbor are objects that are similar to the target in the track space (visually or spatiotemporally) and are the major source of false matches to the target. 
  (a) Tracking without considering neighbors. 
  As shown in the graph on the right, the association problem can be viewed as a bipartite matching problem between candidates and the target.
  The edge weight is the similarity. 
  Matching between the target alone and candidates can be unreliable when there is occlusion. 
  As shown in this example, the target is incorrectly matched with candidate$_2$ since it has a lower similarity to the correct match (candidate$_1$) as a result of occlusion.
  (b) Tracking with considering neighbors. 
  In this example, since the neighbor is not occluded, it can be matched with candidate$_2$ correctly.
  As a result, the target must match candidate$_1$, since matching the same candidate is not permitted. 
  The matching procedure can be formulated as a bipartite matching problem between candidates and the target along with neighbors. 
  It can be seen, in this example, that by including neighbors in the matching process, the accuracy of matching and tracking can be improved.
  (The example is from MOT16~\cite{MOT16}.)
  }
  %The upper figure shows the situation when occlusion happens, the left is the target and the right is the candidate position, and the number on top of the edge is the matching confidence score.
  %Due to occlusion, the target is incorrectly matched to other candidate positions.
  %The lower figure shows the results after applying NeighborTrack.
  %Since neighbors are not occluded, they can be used to assist in matching and greatly increase the success rate.
  \iffalse
  \caption{利用鄰居資訊修正追蹤結果:
  上圖為siamese network被遮擋時的狀況，左邊為目標，右邊為候選位置，每條線上的數字為兩兩匹配的confidence score。由於外觀被遮擋的關係，目標會錯誤的匹配到其他候選位置，下圖則是套用我們的方法之後的結果，由於鄰居未被遮蔽，鄰居與candidate的匹配資訊減少了對目標來說錯誤的位置進而增加目標匹配成功的機會。資料庫為\cite{MOT16}}\fi  
  \label{fig:neighbors_similar}
\end{figure*}

To address the challenges caused by appearance changes, we propose a post-processor named NeighborTrack.
The main idea is to verify the correctness of the tracking result using the spatial-temporal trajectories of the neighbors of the target.
NeighborTrack simultaneously tracks the target and its neighbors.
When the target is occluded, it reduces the number of matching errors by involving neighbors in the matching process.

Consider the example in \figref{neighbors_similar}.
In the current frame $I_t$, there are two candidates for the tracking target. 
Numbers on the edges indicate the similarity between two objects.  
In frame $I_t$, the target object is occluded. 
When only considering appearance cues (\figref{neighbors_similar}(a)), the target is incorrectly matched to the wrong candidate$_2$ because of occlusion.
Aside from visual and spatiotemporal cues, NeighborTrack makes use of neighbors in the previous frame in order to resolve ambiguities caused by occlusions. 
Neighbors are objects that are similar to the target object. 
Often, they are the source of ambiguity. 
By considering neighbors (\figref{neighbors_similar}(b)), we construct a bipartite graph with two sets of nodes: one for the hypothesis of matches (candidates) and one for the source of ambiguity (neighbors) and the target. 
After bipartite matching, since the neighbor is not occluded, it can be correctly matched with candidate$_2$. The target should therefore choose candidate$_1$ (the correct one), even if their similarity remains low due to occlusions. 
\iffalse
By matching candidates with neighbors, we remove the most source of ambiguity and can better match the target, even if it is occluded, thereby improving tracking performance. \fi
%Since neighbors are not occluded, they can be used to assist in matching and greatly increase the success rate.
%The upper figure shows the situation when occlusion happens, the left is the target and the right is the candidate position, and the number on top of the edge is the matching confidence score.
%The lower figure shows the results after applying NeighborTrack.
Note that NeighborTrack uses the spatial-temporal tracklets of the neighbors for more robust similarity estimation.
Our experiments indicate that NeighborTrack effectively mitigates the problems caused by occlusion and increases tracking accuracy.

%The target object becomes occluded at frame $I_{t+1}$.
%For example, suppose $I_t$ and $I_{t+1}$ are two consecutive frames, the upper half of Figure \figref{neighbors_similar} shows an example of occlusion. 
%Because the target is blocked, the confidence score of the real target is lower than that of neighboring objects similar in appearance to the target.
%Under these circumstances, the tracker will pick the wrong target and cause the tracking to fail. 
%To solve the above problem, we simultaneously consider the neighboring objects, define their spatial and temporal relationship, and describe them to increase the semantic level between the target and its neighbors.
%As shown in the lower half of \figref{neighbors_similar}, if a neighbor participates in the matching process, since it is not blocked, its confidence score is high, and, therefore, this makes up for the deficiencies caused by the target being blocked.

We propose NeighborTrack, a post-processor for improving SOT, and have made the following contributions.
\begin{itemize}

\item  We formulate  the association problem as a bipartite matching problem between candidate and neighbor tracklets. Unlike attention-based methods that strengthen single-target appearance features, our method uses neighbor information to help correct wrong tracking when appearance features change.

\item Based on cycle consistency, we calculate Intersection over Union (IoU) between two tracklets as a more robust measure of similarity. Both spatiotemporal and visual cues are considered in the measurement.

\item Our approach is not in conflict with, and it can complement, methods for enhancing appearance features or enhancing object localization, such as DIMP~\cite{DIMP} and the Alpha-refine~\cite{alpharefine}.
The agnostic nature of our method enables it to be combined with most SOT methods.
%including traditional attention-based methods and transformer-based methods of more recent development.

\item Our method does not require additional training data or retraining or fine-tuning, and neighbor information is readily available from the network-generated confidence scores without additional computation.

\item We achieved state-of-the-art 72.25${\%}$ AUC on the LaSOT~\cite{lasotDBLP:journals/corr/abs-1809-07845} dataset. For the GOT-10K\cite{got10kHuang2021} dataset, we achieved 75.7${\%}$ AO. On the bbox and mask datasets of VOT challenge~\cite{VOT_TPAMI}, our method increases EAO and robustness by ${2.11\%}$ and ${1.92\%}$, respectively.

\end{itemize}

%% file: 2related.tex
\section{Related Work}
\label{sec:related}

SOT methods based on Siamese networks~\cite{siameseFCbertinetto2016fullyconvolutional} have been continuously refined and improved.
The majority of research has focused on improving the header function by adding more effective tracking mechanisms.
Some methods~\cite{siamrpnpp_Li_2019_CVPR, siamrpn_Li_2018_CVPR} borrow the region proposal network mechanism from Faster R-CNNs~\cite{ren2016faster} to make a network more adaptive to target size changes.
Wang~\etal~\cite{siammask_wang2019fast} propose a method that generates a mask for tracking the target while training the tracker in order to enable the network to track targets more accurately and interpretably when used in real scenes. 
Ocean~\cite{ocean_zhang2020} replaces anchor-based headers with anchor-free ones to alleviate issues associated with overlapping tracked targets and anchors in a crowded scene. 
Aside from the header issue, effective extraction of appearance features has always been a focus of SOT research.
Recent methods~\cite{TransT,mixformer,ostrack} achieve breakthrough performance by employing a transformer network to simultaneously track the target and all the surrounding backgrounds, and then establishing an appearance feature model covering both self-attention and cross-attention. 

Although the above-mentioned methods improve object tracking, they cannot overcome the tracking difficulties caused by occlusion~\cite{Lee_2014}. 
Siamese network-based methods often fail when appearances change; therefore, relying solely on appearances is not unreliable.
The confidence score predicted by a siamese network is a good indicator of appearance similarity, and objects with a similar appearance to the target will have high confidence scores.
According to the tracking results on VOT challenge~\cite{VOT_TPAMI} reported by Zhang~\etal~\cite{ocean_zhang2020}, when non-tracking objects occlude the target object, the target appearance will change significantly, and this will greatly reduce the confidence score of the target position.
It is therefore necessary to consider other information to overcome the challenge of occlusion.

The use of neighboring information is common in most multi-object tracking (MOT) systems.
Even though the MOT task must consider multiple-to-multiple tracking, it is actually equivalent to taking into account the time-space matching relationship of all objects in the search range at the same time.
There are, however, two differences between SOT and MOT.
(1) Most MOT networks such as FairMOT~\cite{fairmotzhang2021} only track certain pre-determined classes.
In contrast, a general SOT system should be able to track virtually any type of object. 
The feature space of an MOT model can be pre-trained on a class-specific Re-ID task to distinguish objects within a class, which is not applicable to objects outside the class.
To be more specific, MOT has a considerable degree of pertinence in extracting appearance features.
Targets will not be significantly affected by appearance features as long as their key parts are not occluded.
As for SOT, it does not have the above class information, and the requirements for appearance characteristics must be universal.
Therefore, it is more susceptible to occlusion problems.
(2) In the design of an MOT system, objects of the same class are tracked together. 
This is actually beneficial to tracking because the matching relationship between multiple objects and the background or other objects can be easily defined and then used to correct the tracking results.
As for SOT, it does not have direct access to information about object adjacency.
In DIMP~\cite{DIMP}, the authors propose to learn foreground/background information directly so as to adapt to different backgrounds in different videos.
But, this approach essentially only considers the appearance characteristics of a single target, as opposed to an MOT system which directly uses the matching relationship between multiple objects.
Due to the nature of the SOT task, it is not possible to predict in advance the information regarding class and Re-ID.
This paper presents a systematic approach to utilizing neighbor information for SOT.

Some methods~\cite{sotreidZhang_2021_CVPR,keepTrackmayer2021learning} take into account multiple objects in the SOT task.
DMTrack~\cite{sotreidZhang_2021_CVPR} adds a Re-ID network without class information to the SOT task; however, it should be noted that the Re-ID feature cannot be easily generalized to data that have not yet been trained before~\cite{personreiddomainchansferDeng_2018_CVPR}.
KeepTrack~\cite{keepTrackmayer2021learning} trains a graph neural network to determine the tracking result after referring to all candidate regions in two adjacent frames.
In comparison with these methods that utilize information across objects or frames, our method offers the following advantages: 
(1) Our method is a post-processing method. Thus, unlike these methods, our method does not require an additional network and retraining to obtain neighbor information, making it more flexible to use. 
(2) These methods only take into account information from two adjacent frames. Whenever a target is occluded, at least one frame's target appearance features will be unreliable. 
As a result of incorporating past information for a period of time, our method is able to better reduce tracking inaccuracies resulting from occlusions or changes in appearance over time.

%% file: 3method.tex
\section{NeighborTrack}
\label{section:ourmethod}

\begin{figure}[t]
  \centering
  %\fbox{\rule[-.5cm]{4cm}{4cm} \rule[-.5cm]{4cm}{0cm}}
  \includegraphics[width=0.98\linewidth]{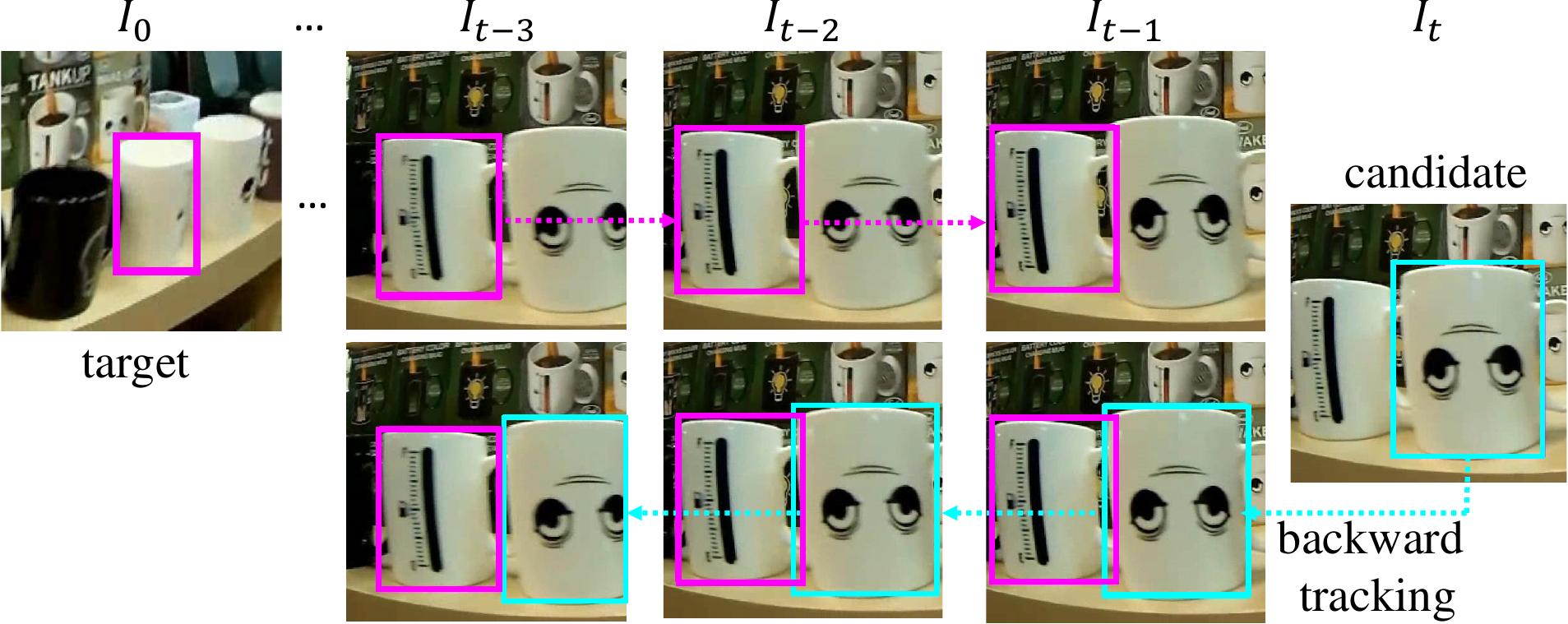}
  \caption{
  A real example of using cycle consistency to eliminate incorrect candidates
  The example is from LaSOT~\cite{lasotDBLP:journals/corr/abs-1809-07845}.
  Based on the target template at $I_0$, the tracker (OSTrack~\cite{ostrack} in this example) tracks it until the frame $I_{t-1}$ and produces a target tracklet (magenta boxes).
  For the current frame $I_t$, it selects the incorrect candidate due to severe view changes from the template.
  As a result of backtracking the candidate, we are able to obtain its backward-tracking candidate tracklet (cyan boxes). 
  After calculating its similarity to the forward-tracking target tracklet, it is clear that the candidate does not match the target well. 
  Using the proposed measure, we are able to verify matches better by utilizing the spatiotemporal relationship. 
  %Cycle consistency gives additional temporal and spatial clues.
  %Here the green bbox is ground truth, and the baseline model is OSTrack\cite{ostrack}.
  %We use the magenta bbox to represent the tracking result, and the cyan bbox to represent reverse tracking answer.
  %After the calculation of the IoU combined with the historical trajectory and cyan bbox, we found that the baseline result does not match the cycle consistency IoU of the historical trajectory.
  %We rely on this time-space relationship clue to assist in the tracking.
  }
  \iffalse\caption{cycle consistency提供額外的時空間線索：綠色bbox為ground tuuth,baseline model為OSTrack\cite{ostrack}。追蹤結果為紫色bbox，藍色bbox為歷史軌跡，每個$CAND$反向追蹤後與歷史軌跡計算IoU可以發現baseline結果與歷史軌跡的cycle consistency IoU並不吻合，這種時空間關係的線索可以幫助追蹤。資料庫為\cite{lasotDBLP:journals/corr/abs-1809-07845}
  Cycle consistency gives additional temporal and spatial clues.
  Here the green bbox is ground truth, and
  the baseline model is OSTrack\cite{ostrack}.
  We use the magenta bbox to represent the tracking result, and
  the blue bbox to represent historical trajectory.
  After the reverse tracking of each $CAND$ and
  the calculation of the IoU combined with the historical trajectory,
  we found that the baseline result does not match
  the cycle consistency IoU of the historical trajectory.
  We rely on this time-space relationship clue to assist in the tracking.
  (The dataset chosen for this example is from\cite{lasotDBLP:journals/corr/abs-1809-07845}.}
  \fi
  \label{fig:cycleconsistency_realcase}
\end{figure}

This paper proposes NeighborTrack for improving the tracking results of an SOT network $\Phi$ as long as it meets two requirements. First, given a target template patch $z$, its bounding box $b_0$ specifying its position at the first frame $I_0$, and a sequence of frames $(I_1,\cdots,I_{T})$, the network $\Phi$ can generate a tracking result $(b_1,\cdots,b_{T})$ where $b_t$ is the bounding box indicating the position of the target object at frame $I_t$.
%, \ie, $\Phi(z, p, \{I_1,\cdots,I_{\tau}\})=\{b_1,\cdots,b_{\tau}\}$. 
Second, for a frame $I_t$, the network can generate a set of candidate bounding boxes $B^t$ and their confidence scores $S^t$. The two requirements are met by the majority of SOT networks.

To determine the bounding box of the target object at the current frame $I_t$, NeighborTrack produces a set of candidate bounding boxes $\mathcal{C}^t$. 
These are the hypotheses that the target matches. 
Among the candidates, we are seeking to identify the best match for the target.
To have a more robust similarity measurement, NeighborTrack is inspired by the principle of cycle consistency~\cite{cycleconsistency}: an object tracked forward along a time line should be able to return to its original position when tracked backwards along time.
In the current frame $I_t$, we have a forward-tracking tracklet $(b_{t-\tau},\cdots,b_{t-1})$ for the target object.
For the purpose of utilizing cycle consistency for similarity, we backtrack each candidate using $\Phi$ for $\tau$ frames in order to extract its backward-tracking tracklet. 
A candidate tracklet pool $\mathbf{P}^c$ is formed by these tracklets.
Similarity between two tracklets is determined by their intersection over union (IoU) value. 
The measurement considers both visual and spatiotemporal cues since IoU takes the spatiotemporal cue into account, while visual cues are considered when tracklets are formed through forward/backward tracking. 
By measuring the similarity between the forward-tracking target tracklet and the backward-tracking candidate tracklet, we can verify how closely a candidate matches the target object using forward/backward tracking and cycle consistency.
\figref{cycleconsistency_realcase} illustrates a real example of how a tracker can eliminate incorrect matches by ensuring cycle consistency.

NeighborTrack also maintains a neighbor tracklet pool $\mathbf{P}^n$ which contains neighbors that are similar to the target.
Since neighbors are similar to the target (visually and spatiotemporally), they can cause false matches, particularly when the target's appearance changes or when it is obscured.
As illustrated in \secref{intro}, the neighbors are used to resolve ambiguity. % when occlusions or appearance changes occur.

%\secref{CANDandtrajectorypool} describes the procedures for obtaining the candidate set, the candidate tracklet pool, as well as the neighbor tracklet pool. 
%\secref{associations} explains how to correlate the target tracklet with the candidate tracklets through bipartite matching. 

%NeighborTrack aids the tracking task of a siamese SOT network.
%It maintains a candidate set ($CAND$) and a trajectory pool.
%$CAND$ holds the candidate positions of the tracking results, and the trajectory pool records historical trajectories of the target as well as the trajectories of its neighbors.
%Since $CAND$ may contains non-target objects, we introduce neighbor information through the association between
%$CAND$ and trajectory pool taking inspiration from cycle consistency\cite{cycleconsistency}, \ie, an object that is tracked forward along the time axis should be able to return to its original position if it is tracked backwards along the same axis.
%Based on the above hypothesis, we rely on the Intersection over Union (IoU) of forward/backward tracking to verify whether $CAND$ is related to the trajectory in the trajectory pool, and then use this information to correct the wrong tracking result.
%As shown in \figref{cycleconsistency_realcase}, the concept of cycle consistency can provide timing clues to help us target correctly.
% We will describe how to form $CAND$ and maintain the trajectory pool in Section \ref{sec:CANDandtrajectorypool}, and how to correlate $CAND$ and trajectory pools with each other in Section \ref{sec:associations}.
%}

\subsection{Candidate and neighbor tracklets}\label{sec:CANDandtrajectorypool}

%Next we describe $CAND$ and trajectory pooling in detail.
\heading{Candidate set.} As one of the requirements, given the previous tracking results $(b_1,\cdots,b_{t-1})$ for $(I_1,\cdots,I_{t-1})$, the SOT network can generate a list of $n_t$ candidate bounding boxes $\tilde{B}^t=\{\tilde{b}^t_1,\cdots,\tilde{b}^t_{n_t}\}$ and their corresponding confidence scores $\tilde{S}^t=\{\tilde{s}^t_1,\cdots,\tilde{s}^t_{n_t}\}$ for the current frame $I_t$. Most SOT methods find the bounding box $\tilde{b}^t_{i_{max}}$ with the highest confidence score as the tracking result of $I_t$, where $i_{max} = \arg\max_{i} \tilde{s}^t_{i}$. Instead of picking up the most confident one, NeighborTrack maintains a candidate set of bounding boxes $\mathcal{C}^t$ and finds the match within $\mathcal{C}^t$. We first filter out bounding boxes with insufficient confidence scores:
\begin{align}
(\hat{B}^t, \hat{S}^t) = \{ (\tilde{b}^t_i, \tilde{s}^t_i) \: | \: \tilde{b}^t_i \in \tilde{B}^t \mbox{ and } \tilde{s}^t_i > \alpha \tilde{s}^t_{i_{max}})\},
\end{align}
where $\alpha \in [0, 1]$ is a hyperparameter for the threshold confidence ratio, $\hat{B}^t$ is the set of candidate bounding boxes with sufficient confidence and $\hat{S}^t$ are their confidence scores. 
Next, we perform non-maximum suppression on the remaining bounding boxes:
\begin{align}
(B^t, S^t) = \mbox{SoftNMS}(\hat{B}^t, \hat{S}^t),
\end{align}
where $\mbox{SoftNMS}$ is an improved version of non-maximum suppression~\cite{softnms}, $B^t$ and $S^t$ respectively contain the bounding boxes and adjusted scores after $\mbox{SoftNMS}$.

If the target object is severely occluded, the correct match may not be included in the candidate set even if a loose threshold is applied.
Thus, we also apply a Kalman filter to predict the candidate bounding box $b^{\kappa}$ without relying on appearance features.
The candidate set $\mathcal{C}^t$ is formed by adding $b^{\kappa}$ into $B^t$, \ie, $\mathcal{C}^t = B^t \cup \{ b^{\kappa} \}$.

%We know that kalman filter may overlap with other $CAND$, but kalman filter and other $CAND$ are obtained by different methods, so kalman filter should not calculate SoftNMS with other $CAND$.

%When the target is completely occluded, it may not appear in the $CAND$ despite the looser appearance features used as the condition for $CAND$.
%To prevent this from happening, we apply a Kalman filter to generate a reference position independent of appearance features, and add it to $CAND$. We know that kalman filter may overlap with other $CAND$, but kalman filter and other $CAND$ are obtained by different methods, so kalman filter should not calculate SoftNMS with other $CAND$.}
 %提前把kalman filter的敘述加到這裡
 
\heading{Candidate tracklet pool.} For each candidate $b^t_i \in \mathcal{C}^t$, we generate its tracklet $\xi^t_i$ by backtracking for $\tau$ frames.
We set the patch $z^t_i$ as the target template which is the patch cropped from using the bounding box $b^t_i$ and use the SOT network $\Phi$ for backtracking, \ie, 
\begin{align}
\xi^t_i = \Phi(z^t_i, b^t_i, (I_{t-1},\cdots,I_{t-\tau})).  
\end{align}
The tracklet $\xi^t_i$ is a sequence of bounding boxes indicating the positions of the target template $z^t_i$ from time $t\!-\!1$ to time $t\!-!\tau$.
The set of all candidate tracklets is referred to as the candidate tracklet pool $\mathbf{P}^c$.
Intuitively, the candidate tracklet pool contains the backtracking tracklets of objects that could potentially be the target object.

%\bm{$CAND$}: As shown in Figure \ref{fig:confidence score and CANDidate}, we use a siamese network called S-network as the base network.
%For each frame, the S-network generates a confidence score $C_i$ and a bounding box (bbox) $B_i$ for each anchor position $i \in\{1, 2,\dots, wh\}$, where $wh$ is the total number of $C$, in the search region.
%Usually, $C_j$ and $B_j$ corresponding to the position $j$ (where $j=\arg\max_{i} C_i$) with the highest confidence score will be chosen as the tracking result.
%In NeighborTrack, we first establish a dynamic threshold $\alpha C_j$, where $\alpha \in [0, 1]$ is a hyperparameter.
%We use $\alpha C_j$ to filter out those low-confidence scores and bboxes, and then select high-quality tracking results from those remaining, as shown in the following equation:
%\begin{equation}\label{eq:CAND}
%CAND = \{SoftNMS(B_i| C_i > \alpha C_j) %\}. 
%\end{equation}

%The SoftNMS function is adopted from \cite{softnms}.
%To correlate $CAND$ with the trajectory pool, we input each member of $CAND$ as a tracking kernel into the S-network as shown in Figure \ref{fig:inverse_track}.
%Let $n$ be the time duration for backtracking, then $CAND$ at time $t$ will back track from\ the bbox frame $I_{t-1}$ to that of frame $I_{t-n}$.
%We refer to $CAND$ and its backtracking result together as the $CAND$ trajectory (from frame $I_{t}$ to frame $I_{t-n}$).

\heading{Neighbor tracklet pool.} NeighborTrack also maintains another tracklet pool called the neighbor pool $\mathbf{P}^n$. 
Neighbor tracklets are essentially the candidate tracklets from the previous frame. 
By applying the method described in section \secref{associations}, NeighborTrack selects a bounding box $b^t_m$ in the candidate set $\mathcal{C}^t$ as the tracking result for $I_t$.
Afterwards, NeighborTrack updates the neighbor pool $\mathbf{P}^n$ using the current candidate tracklet pool $\mathbf{P}^c$.
In the first step, the selected tracklet $\xi^t_m$ is removed from $\mathbf{P}^c$.
For each tracklet $\xi^t_i$ remaining in $\mathbf{P}^c$, we adjust its time span from $[t\!-\!1, t\!-\!\tau]$ to $[t, t\!-\!\tau\!+\!1]$ to be ready for the next frame $I_{t+1}$.
This is accomplished by appending the associate candidate bounding box $b^t_i$ at the head and removing the last bounding box:
\begin{align}
\zeta^{t+1}_i = (b^t_i)^\frown \rho(\xi^t_i),
\end{align}
where $^\frown$ is the concatenation operator of two sequences and $\rho(s)$ removes the last element from the sequence $s$.
The neighbor tracklet $\mathbf{P}^n$ is updated by $\{ \zeta^{t+1}_i | i \neq m \}$ for the next frame $I_{t+1}$.
As a result, the neighbor tracklets are the unselected candidate tracklets from the previous frame (after aligning the time span to the current frame).
%\torevise{Initially, the neighbor pool is empty when the tracking is initiated. A neighbor pool is gradually built throughout the tracking process. The time span gradually increases to $\tau$ after tracking for $\tau$ frames and remains $\tau$ afterwards.}

We maintain the neighbor tracklets because they belong to neighbors of the target object in the tracking space. 
They are similar to the target object either visually or spatiotemporally and may cause ambiguity to the SOT network, particularly when the target object is obscured.
By bipartite matching with those neighbors (\secref{associations}), the ambiguity could be better resolved. 

We only retain neighbor tracklets from the previous frame, not those from earlier frames.
There are several reasons for this.
%Firstly, older tracklets often have difficulty finding correspondences. 
To begin with, tracking accuracy degrades over time as outdated neighbor tracklets rarely provide useful information.
Additionally, if a neighbor tracklet continues to survive, it should be possible to find the correspondence at the most recent time.
Finally, retaining more tracklets will increase computation overhead.
Thus, to avoid filling the pool with outdated neighbor tracklets that slow down computation speed, we only retain candidate tracklets from the previous time instance.

\ignore{
\todo{add \figref{inverse_track}?}
\torevise{The gray bbox at frame $I_t$ is $CAND$,
and the blue bboxes from frame $I_{t-1}$ to $I_{t-n}$
together with the orange dashed bboxes form the tracklet pool.
The blue bbox is the target historical trajectory,
and the orange dashed bbox is the neighbor trajectory.
It should be noted that the neighbor trajectories in the trajectory pool
are in $CAND$ previously
that failed to match at the previous time point,
and we do not keep distant neighbor trajectories.}
}

\subsection{Tracking by bipartite matching}
\label{sec:associations}

Through the process introduced in \secref{CANDandtrajectorypool}, we have the candidate pool $\mathbf{P}^c$ and the neighbor pool $\mathbf{P}^n$. 
The candidate pool contains hypothesis of tracking results while the neighbor pool contains the source of potential ambiguity.
Given the target tracklet $\eta=(b_{t-1},\cdots,b_{t-\tau})$ which is the tracking results for the previous $\tau$ frames, our goal is to find the association of $\eta$ among hypothesis $\mathbf{P}^c$ while verifying with the source of ambiguity $\mathbf{P}^n$. 

\begin{figure}[t]
  \centering
  %\fbox{\rule[-.5cm]{4cm}{4cm} \rule[-.5cm]{4cm}{0cm}}
  \includegraphics[width=0.95\linewidth]{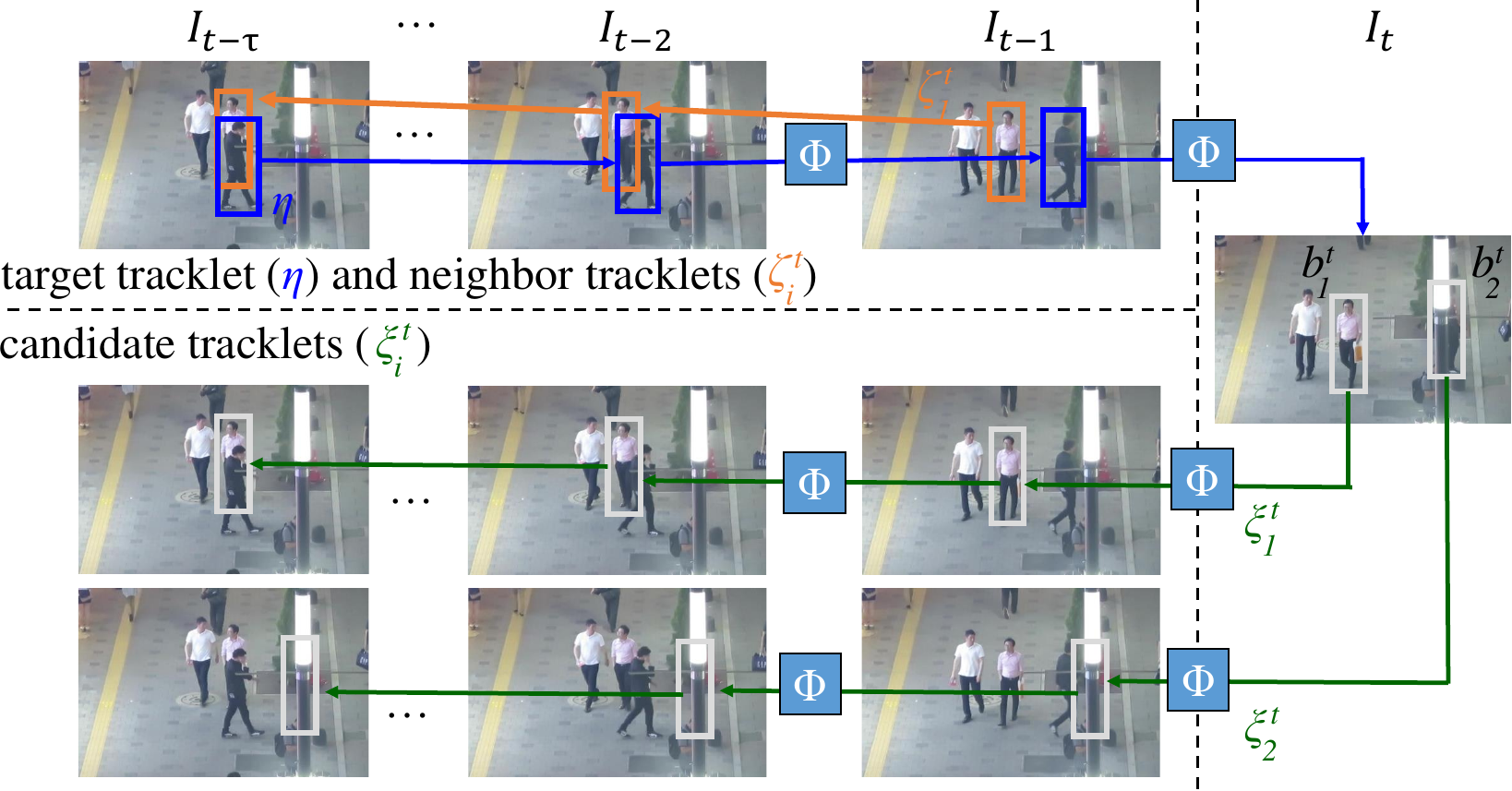}
  \caption{
  An illustration of the candidate and neighbor tracklets. 
  The blue boxes represent the target tracklet ($\eta$), while the orange boxes represent a neighbor tracklet ($\zeta^t_1$). For the current frame $I_t$, the tracker $\Phi$ generates a set of candidates (the white boxes), $b^t_1$ and $b^t_2$, in $I_t$.
  Using $\Phi$ for backtracking, we obtain their candidate tracklets ($\xi^t_1$ and $\xi^t_2$).
  Note that since the target object is occluded in $I_t$, the confidence score of the correct candidate ($b^t_2$) is lower than that of the incorrect candidate ($b^t_1$). 
  However, as $b^t_1$'s candidate tracklet ($\xi^t_1$) matches very well with the neighbor tracklet ($\zeta^t_1$), bipartite matching decides to match it with the neighbor rather than the target.
  Therefore, the target $\eta$ has to be matched with the candidate $\xi^t_2$, thus correcting the false tracking. 
  The example is from MOT16~\cite{MOT16}.  
%  Reverse tracking of historical trajectory and $CAND$: the blue box represents the historical trajectory, the orange dotted line is the neighbor trajectory, and each frame $I_t$ passing through the S-network will generate $CAND$ (gray bbox).
%  Each $CAND$ obtained from the above will use the S-network to track the trajectory of $CAND$ in reverse along the time axis.
%  The forward tracking kernel is provided by the ground truth, and the reverse tracking kernel is generated by $CAND$.
%  The yellow arrow denotes the direction of the data flow across multiple time periods.
%  These dashed yellow arrows move forward or backward without skipping any points in time.
%  (The dataset is from\cite{MOT16}.)
  }
  \iffalse\caption{歷史軌跡與$CAND$的反向追蹤:藍色框為目標歷史軌跡，虛線的橘色框為鄰居軌跡，而每個經過S網路的frame $I_{t}$會產生$CAND$ (灰色框)，每個$CAND$會分別利用S網路沿著時間軸反向追蹤得到$CAND$軌跡。正向追蹤的追蹤核由GT提供，反向追蹤的追蹤核則由$CAND$產生。黃色箭頭為資料流的方向，為了簡略表示，跨越多個時間的資料流為虛線黃箭頭。其依然沿著時間順序向前或向後，並不會跳過任何時間點。資料庫為\cite{MOT16}}\fi
  \label{fig:inverse_track}
\end{figure}

We cast the association problem into a bipartite matching problem. 
We have two sets of tracklets: $\mathbf{S}^c = \mathbf{P}^c$ and $\mathbf{S}^n = \mathbf{P}^n \cup \{\eta\}$.
Each tracklet in $\mathbf{S}^c$ and $\mathbf{S}^n$ can be taken as a node. 
%Assume there are $N_c$ tracklets in $\Xi^c$ and $N_n$ tracklets in $\Xi^n$.
Thus, we have a bipartite graph whose two independent sets of nodes formed by $\mathbf{S}^c$ and $\mathbf{S}^n $.
It is a complete bipartite graph since every node of the first set $\mathbf{S}^c$ is connected to every node of the second set $\mathbf{S}^n$. 
The weight $w_{ij}$ associated with the edge between two nodes, $\xi^t_i \in \mathbf{S}^c$ and $\zeta^t_j \in \mathbf{S}^n$, is defined as the average IoU values between two tracklets. That is, if $\xi^t_i=( b^c_{t-1},\cdots,b^c_{t-\tau})$ and $\zeta^t_j=( b^n_{t-1},\cdots,b^n_{t-\tau})$, then we have
\begin{align}
w_{ij} = \frac{1}{\tau} \sum_{k=t-\tau}^{t-1} \mbox{IoU}( b^c_k, b^n_k),
\end{align}
where $\mbox{IoU}$ calculates the IoU values between two bounding boxes. 
The weight reflects the trajectory similarity between two tracklets. 
We employ the Hungarian algorithm~\cite{hungarian_Munkres1957} to find the maximum matching for the resultant bipartite graph.
Suppose that the candidate tracklet $\xi^t_m$ is paired with the target tracklet $\eta$, its corresponding candidate bounding box $b^t_m$ is selected as the tracking result for frame $I_t$.
When the target tracklet $\eta$ is not matched, we select the non-matched candidate tracklet with the highest IoU values with $\eta$. 
If the highest IoU is zero, we select the candidate $b^{\kappa}$ predicted by the Kalman filter.
%\todo{what if the target tracklet does not have a match?}

It is important to note that, although we only take the match of the target tracklet as the tracking result, all neighboring tracklets serve as a means of verifying the tracking result.
If a candidate tracklet has a high IoU similarity to the target tracklet, it is likely to be the target. 
In contrast, if it has a high IoU similarity with any neighbor tracklet, it is likely to be associated with another object rather than the target. In this way, neighbor tracklets contribute to the matching process and assist in resolving ambiguities. 

Tracking is significantly hampered by occlusions. 
Since the appearance feature is unreliable, it is easy to lose track of a target that has been occluded in the current frame. 
NeighborTrack addresses this issue with the help of neighbor tracklets. 
Assuming that the target object is occluded in the current frame, the correct candidate tracklet will have a lower similarity to the target tracklet. 
However, NeighborTrack can still find the correct tracking result as long as the objects in the candidate set are not occluded.
Since the candidate objects are not occluded, their candidate tracklets can find excellent matches among the neighbor tracklets and will not select the target tracklet as their matches.
\figref{inverse_track} gives an example on how bipartite matching works.

Our edge weight only considers the similarity of bounding boxes, and not the visual similarity of patches. 
There are two reasons for this. First of all, the appearance feature of an occluded target is less unreliable. In the event that the target is occluded, its appearance may resemble that of the occluder. Thus, visual similarity could mislead the results. 
Secondly, we construct candidate tracklets by backtracking patches of the candidate objects. In this way, the appearance features of candidates have already been implicitly considered in the tracklets.

\ignore{
\begin{figure*}
  \centering
  %\fbox{\rule[-.5cm]{4cm}{4cm} \rule[-.5cm]{4cm}{0cm}}
  \includegraphics[width=0.95\textwidth]{target_in_neighborframes_v14.png}
  \caption{The relationship between confidence score and $CAND$.
  Given target and search, the siamese network will generate
  $wh$ confidence scores $C_i , i\in[1,wh]$,
  within the search range.
  Here the confidence score $C_i$ denotes
  the possibility of the target falling at that position,
  and each $C_i$ will get the corresponding bbox after header processing.
  We call the bbox obtained above $B_i$, and
  the filtered $B_i$ will be marked as a gray bbox and called $CAND$.
  (The dataset chosen for this example is from\cite{MOT16}.)
  }
  \iffalse\caption{confidence score and $CAND$之間的關係。在Siamese Network輸入Target與Search後，系統會在Search範圍內產生wh個Confidence score， $C_i , i\in[1,wh] $。在這裡每個點的confidence score $C_i$代表目標落在該位置的可能性，而每個$C_i$經過Header處理後會得到對應的bbox。我們稱上述得到的bbox為$B_i$，而經過篩選後的$B_i$會被標記為灰色bbox並稱為$CAND$。資料庫為\cite{MOT16}}\fi
  \label{fig:confidence score and CANDidate}
\end{figure*}

\begin{figure*}[t]
  \centering
  %\fbox{\rule[-.5cm]{4cm}{4cm} \rule[-.5cm]{4cm}{0cm}}
  \includegraphics[width=0.73\textwidth]{bipartite_matching_v6.png}
  \caption{This figure illustrates the Bipartite matching between K1 ($CAND$) and K2 (trajectory pool). Each element in $CAND$ and the trajectory pool is treated as a node to execute Bipartite matching. The edge between nodes is the average IoU connecting the two trajectories, and the bbox below is a schematic diagram of the two trajectories corresponding to the edge. (The dataset chosen for this example is from\cite{MOT16}.)
  }
  \iffalse\caption{Bipartite Matching between $CAND$ ($K1$)  \&\ 軌跡池 ($K2$):將$CAND$軌跡與軌跡池的每個元素各自當成一個節點做Bipartite matching。節點之間的邊為各自軌跡的平均IoU，而下方的bbox為該條邊對應的兩條軌跡之示意圖。資料庫為\cite{MOT16}}\fi
  \label{fig:bipartite_matching}
\end{figure*}

\begin{figure*}[t]
  \centering
  %\fbox{\rule[-.5cm]{4cm}{4cm} \rule[-.5cm]{4cm}{0cm}}
  \includegraphics[width=0.95\textwidth]{get_tracking_answer_v4.png}
  \caption{This figure shows how the trajectory pool and tracking results are updated by bipartite matching. In Figure \ref{fig:bipartite_matching}, bipartite matching will determine the best match of the target tracklet, and the corresponding $CAND$ will become the tracking result of Frame $I_t$. The unmatched trajectories become new neighbor trajectories and are used to update the trajectory pool and clear old neighbor trajectories.
  }
  \iffalse 
  \caption{藉由Bipartite Matching更新追蹤結果與軌跡池：Bipartite Matching(圖\ref{fig:bipartite_matching})會決定目標歷史軌跡的最佳匹配，並且對應的$CAND$會成為frame $I_t$的追蹤結果。至於未匹配軌跡則會成為新的鄰居軌跡並被用來更新軌跡池及清除舊的鄰居軌跡。資料庫為\cite{MOT16}}
  \fi
  \label{fig:get_tracking_answer}
\end{figure*}
}

%代表邊的權重，cycle consistency的概念起源自\cite{cycleGAN}，其主要原理是利用資料由函數空間a轉換到函數空間b後再轉換回a時應該等於原本資料的特性作為自監督訓練的損失函數。於2019年由\cite{cycleconsistency}等人推廣到自監督的SOT網路作為損失函數，原理是時間軸正序播放與倒序播放不應該影響追蹤的結果，兩節點的邊定義為節點間對齊時間軸後計算同時間的bbox IoU後平均,視覺化例子如圖\ref{fig:nodeandedge}所述，屬於相鄰時間$\hat{N}^{t-1}$與$\hat{N^t}$的每個節點都將建立一條邊；即式\eqref{eq:edge}，兩相臨時間的節點與邊可以形成成本矩陣$M(\hat{N}^{t-1},\hat{N^t})$:

%每個節點$N^t_{k}$會經由\ref{section:associations}節產生一組對應著由時間段t-$inv_{t}$到t的bbox trajectory資訊$Hist,H_{k}$，$inv_{t}$代表每一個節點總共要紀錄多少frame的歷史位置。當某個節點被選為$N^t_{ans}$的時候會將當下的位置更新在追蹤答案$H_{ans}$中，注意，僅有$H_{ans}$的內容是由紀錄並更新每個時間點的追蹤結果得到的，除了$H_{ans}$以外的所有$H_k$都是由\ref{section:associations}節的式\eqref{eq:model}\eqref{eq:hist}回推得到的結果。

%$\mathop{\sum_{i=1}^{n}}_{i\neq j} a_i$

%% file: 4experiments.tex
\section{Experiments}
\label{sec:experiments}

We begin by discussing the details of implementation in this section.
Then, we apply NeighborTrack to improve several SOT networks and report the results on the VOT~\cite{VOT_TPAMI}, LaSOT~\cite{lasotDBLP:journals/corr/abs-1809-07845} and GOT-10K~\cite{got10kHuang2021} datasets. 

%而在dataset方面，一些比較有名的dataset如imagenet\cite{vid_ILSVRC15}的video image detection、got-10K\cite{Huang_2019}和提供物件mask資訊的ytbvos\cite{xu2018youtubevos}、coco\cite{lin2015microsoft}都是普遍被大家使用的資料集。由於SOT本身需要能夠追蹤各種物件，許多方法如\cite{siamrpn_Li_2018_CVPR}\cite{siammask_wang2019fast}\cite{ocean_zhang2020}用來訓練的資料庫都不只一種，並且在參加vot challenge\cite{VOT_TPAMI}時會利用其訓練集微調(tune)超參數以適應該資料庫。如\cite{siammask_wang2019fast}\cite{ocean_zhang2020}等方法的訓練流程是使用複數的資料庫訓練並且僅在vot challenge\cite{VOT_TPAMI}上tune超參數。

\subsection{Implementation details}

We used the following parameters in all experiments.
A threshold ratio of $\alpha = 0.7$ is used to select candidate bounding boxes. 
%A neighbor with a confidence score larger than $C_j\times 0.7$ is included in $CAND$.
For SoftNMS~\cite{softnms}, the IoU threshold is set to 0.25, and $\sigma$ is set to 0.01 for the Gaussian penalty function.
The time period $\tau$ of backtracking tracklets is 9.
In order to maintain neighbor information and enforce bipartite matching, our method slightly lowers the frame rate, as shown in Table \ref{tabular:time}. The hardware used in the experiment is 8 GTX 1080TI and 2 Intel E5 2620v4 CPU. When the time period $\tau$ is equal to 9, the frame rate is 43\% of the baseline. Also, not including Kalman filter will slightly reduce the performance of our method $( 0.722 \rightarrow 0.720 $)  when setting the same $\tau$. As for the frame rate, it will increase slightly  $( 1.58 \rightarrow 1.71 $) due to the number decrease of the neighbors. 

\iffalse
\todo{為了維護鄰居訊息以及執行二分匹配，我們的方法會使FPS降低，如表\ref{tabular:time}所示，電腦硬體使用8張GTX1080ti與兩顆intel E5 2620v4CPU。time period $\tau$越大時FPS會下降的更多，$\tau$等於9的時候FPS是baseline的43\%。另外，作為參考，在設定相同$\tau$時未加入Kalman filter會稍微降低我們方法的效果(0.722->0.720)並且因為鄰居減少所以使FPS稍微上升(1.58->1.71)。}
We measure the tracking time of on a single 2080Ti GPU.
Our method increases the tracking time by an average of 33${\%}$ for the effort of maintaining neighbor information and performing bipartite matching. 
\fi

\begin{table}
\caption{The effect of changing $\tau$ on the calculation speed/AUC: Take for example the experiments on NeighborTrack on LaSOT benchmark\cite{lasotDBLP:journals/corr/abs-1809-07845} using OSTrack as the basis. From the table, we can see that the frame rate decays as $\tau$ increases. In addition, the tracking effect is slightly reduced if Kalman filter is not introduced, but at the same time, the frame rate is slightly improved.
\iffalse
改變$\tau$對於運算速度/AUC/FPS的影響：以使用OSTrack為基底的NeighborTrack在LaSOT benchmarks~\cite{VOT_TPAMI}。 可以看到隨著time period $\tau$增加時FPS隨之衰減。 另外，沒有Kalman filter會造成追蹤效果稍微降低並且稍微提昇FPS。
\fi
}
\centering
\scalebox{0.8}{
\begin{tabular}{ccccc}

\toprule  %添加表格头部粗线
&&LaSOT&\\

model name&AUC$\uparrow$& FPS(Hz)$\uparrow$\\

\midrule  %添加表格中横线

OSTrack384\cite{ostrack}&0.711& 3.63\\
OSTrack384\cite{ostrack}+ours  $\tau$=3&0.714 &2.40\\
OSTrack384\cite{ostrack}+ours  $\tau$=9&0.722 &1.58\\
OSTrack384\cite{ostrack}+ours w/o Kalman filter  $\tau$=9&0.720 &1.71\\
OSTrack384\cite{ostrack}+ours  $\tau$=27&0.720 &0.75\\
OSTrack384\cite{ostrack}+ours  $\tau$=36&0.721 &0.6\\

\bottomrule %添加表格底部粗线

\end{tabular}}
\label{tabular:time}
\vspace{-6mm}
\end{table}

Because NeighborTrack requires extra computation, it is only activated when tracking results become unstable.
We consider tracking results stable and will not activate NeighborTrack if two conditions are met. 
(1) If there is only one candidate in the set $\mathcal{C}^t$, there is no other option to match except $b^t_{i_{max}}$.
(2) If the average IoU between the target tracklet $\eta$ and the most confident candidate tracklet $\xi^t_{i_{max}}$ is higher than a threshold, then the tracking result is stable. 

%Our proposed method requires additional computing to verify the tracking correctness.
%NeighborTrack is activated when the tracking results become unstable.
%We evaluate the tracking results in two ways:
%(1) When the average IoU of the $CAND$ trajectory corresponding to $B_j$ and the target historical trajectory is higher than a threshold $\alpha C_j$(set $\alpha$ to 0.4 here), it means that the reverse tracking and the forward tracking pretty much match each other, and the system will not be triggered at this time; and 
%(2) When only the $CAND$ trajectories from $C_j$ and $B_j$ pass the threshold of $\alpha C_j$, it means that the S-network is very sure of the tracking result, and NeighborTrack will not be activated.

%(2) When there is only one trajectory in $CAND$, it means that the S-network is very sure of the tracking results, and NeighborTrack will not be activated. 
%When the target is completely occluded, it may not appear in the $CAND$ despite the looser appearance features used as the condition for $CAND$.
%To prevent this from happening, we apply a Kalman filter to generate a reference position independent of appearance features, and add it to $CAND$.
%To save computation, if the $CAND$ has only $B_j$ and Kalman-filter-generated results, the system will not be activated.

\begin{figure*}
  \centering
  %\fbox{\rule[-.5cm]{4cm}{4cm} \rule[-.5cm]{4cm}{0cm}}
  \includegraphics[scale=0.47]{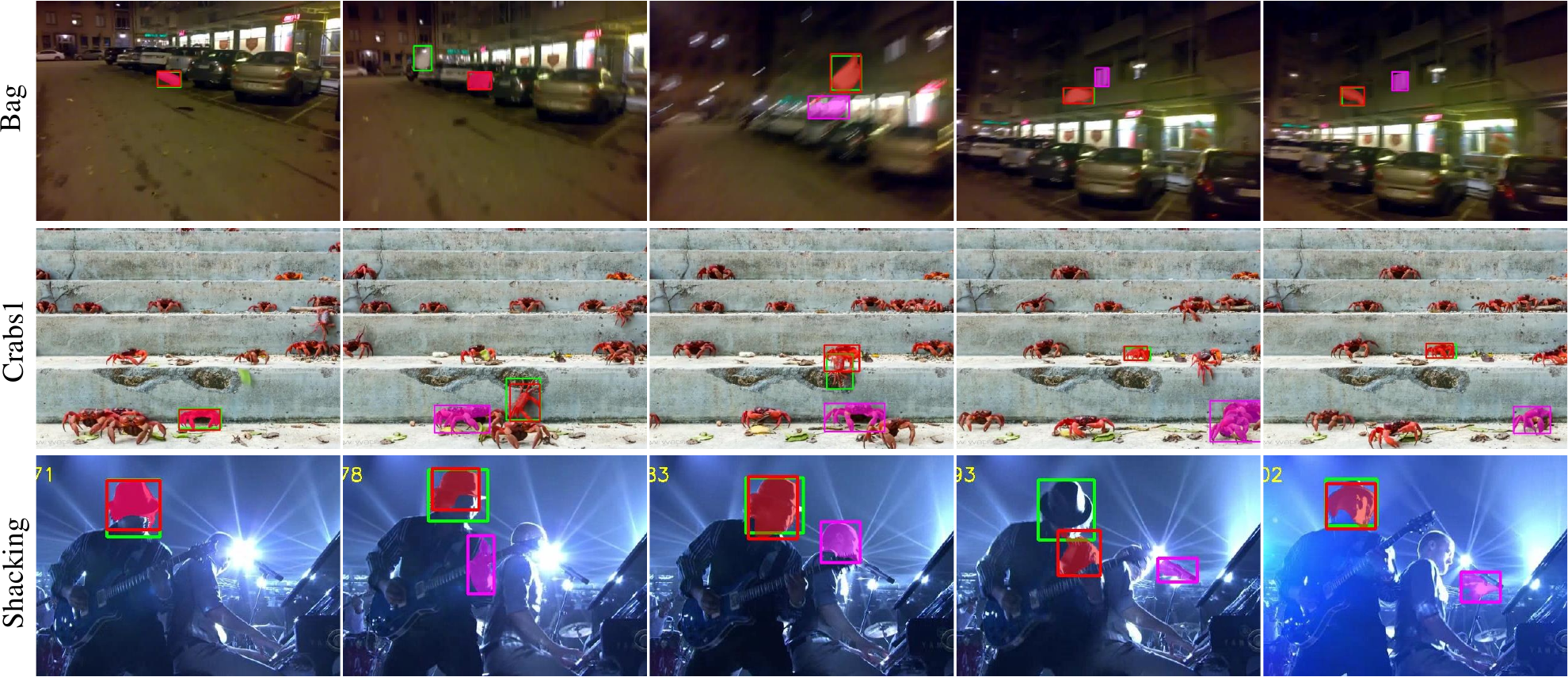}
  \caption{
    Examples of tracking results with and without NeighborTrack from VOT~\cite{VOT_TPAMI}. 
    We use TransT~\cite{TransT} as a baseline and correct its results using NeighborTrack.
    Ground truth is indicated by the green boxes.
    Magenta boxes and masks represent baseline results, while red boxes and masks represent results after applying NeighborTrack.
    Bag and Crabs1 are examples that the baseline is unable to track because there are other objects that resemble the target. 
    In the case of shacking, occlusion and deformation cause severe appearance changes, leading to the failure of the baseline.
    In most cases, tracking errors are corrected after NeighborTrack has been applied.
  }
  \iffalse
  \caption{利用NrighborTrack修正結果:baseline model為TransT\cite{TransT}，綠色bbox為GT，紫色bbox與紫色mask為原始結果，紅色bbox與紅色mask為NrighborTrack修正後的結果。Bag與Crabs1為目標外觀與其他物件相似導致的追蹤困難，Shacking則是遮蔽造成外觀改變導致追蹤失敗。經過我們的方法後都有所改善。資料庫為\cite{VOT_TPAMI}}
  \fi
  \label{fig:votexamples}
\end{figure*}

\begin{table}
\caption{Results of applying NeighborTrack to three different SOT methods on the VOT benchmarks~\cite{VOT_TPAMI}. Green numbers indicate that the original model has been improved.}
\centering
\scalebox{0.8}{
\begin{tabular}{cccc}

\toprule  %添加表格头部粗线
&&VOT2020&\\

model name& accuracy$\uparrow$&	 robustness$\uparrow$& EAO$\uparrow$	\\

\midrule  %添加表格中横线

Ocean\cite{ocean_zhang2020}+AR\cite{alpharefine} 	& 0.757& 0.810& 0.515\\
\color{cyan}Ocean\cite{ocean_zhang2020}+AR\cite{alpharefine}  +ours 				& 0.755& 0.807& \color{green} 0.516\\
%TransT\cite{TransT}+AR\cite{alpharefine} 				& 0.772& 0.761& 0.475\\
%TransT\cite{TransT}+AR\cite{alpharefine}  +ours 			& 0.770& \color{green}0.797& \color{green}0.506\\
TransT\cite{TransT}+AR\cite{alpharefine} 				& 0.769& 0.775& 0.490\\
\color{cyan}TransT\cite{TransT}+AR\cite{alpharefine}  +ours 			& 0.768& \color{green}0.816& \color{green}0.523\\
OSTrack\cite{ostrack}+AR\cite{alpharefine} 			& 0.770& 0.812 & 0.526\\
\color{cyan}OSTrack\cite{ostrack}+AR\cite{alpharefine}  +ours 			& 0.769& \color{green}0.844& \color{green}0.553\\

\midrule  %添加表格中横线

&&VOT2021&\\
\midrule  %添加表格中横线

Ocean\cite{ocean_zhang2020}+AR\cite{alpharefine} 	& 0.756& 0.803& 0.510\\
\color{cyan}Ocean\cite{ocean_zhang2020}+AR\cite{alpharefine}  +ours 				& 0.754& 0.803& \color{green}0.515\\

%TransT\cite{TransT}+AR\cite{alpharefine} 				& 0.771& 0.755& 0.470\\
%TransT\cite{TransT}+AR\cite{alpharefine}  +ours 			& 0.769& \color{green}0.791& \color{green}0.502\\
TransT\cite{TransT}+AR\cite{alpharefine} 				& 0.768& 0.775& 0.494\\
\color{cyan}TransT\cite{TransT}+AR\cite{alpharefine}  +ours 			& 0.767& \color{green}0.809& \color{green}0.519\\
OSTrack\cite{ostrack}+AR\cite{alpharefine} 			& 0.770& 0.810 & 0.528\\
\color{cyan}OSTrack\cite{ostrack}+AR\cite{alpharefine}  +ours 			& 0.769& \color{green}0.843& \color{green}0.556\\

\midrule  %添加表格中横线

&&VOT2022bbox&\\
\midrule  %添加表格中横线

Ocean\cite{ocean_zhang2020} & 0.703& 0.823& 0.484\\
\color{cyan}Ocean\cite{ocean_zhang2020}  +ours 					& 0.703&  0.822& \color{green} 0.486\\

%TransT\cite{TransT} & 0.783& 0.757& 0.476\\
%TransT\cite{TransT}  +ours 			& 0.781& \color{green} 0.800& \color{green} 0.513\\
TransT\cite{TransT} & 0.780& 0.775& 0.493\\
\color{cyan}TransT\cite{TransT}  +ours 			& \color{green} 0.781& \color{green} 0.808& \color{green} 0.519\\
OSTrack\cite{ostrack} 			& 0.779& 0.824& 0.538\\
\color{cyan}OSTrack\cite{ostrack}  +ours 			& 0.779& \color{green}0.845& \color{green}0.564\\

\midrule  %添加表格中横线

&& Average&\\
\midrule  %添加表格中横线
\color{cyan}+ours 						& ${-0.07\%}$& \color{green} ${+2.11\%}$& \color{green} ${+1.92\%}$\\

\bottomrule %添加表格底部粗线

\end{tabular}}
\label{tabular:vot}
\end{table}

\subsection{Datasets and competing methods} 
\label{sec:datasets}

%1.實驗設計
%    1.1短時間dataset證明我們方法對遮擋與外觀改變的有效性
%        用vot證明通用性
%        
%    1.2長時間dataset證明我們方法的穩定性
%        用另兩個證明優越性

We conduct experiments on both short-term and long-term tracking.
For short-term tracking, we use the VOT challenge~\cite{VOT_TPAMI} datasets including VOT2020, VOT2021, and VOT2022bbox~\cite{vot2020Matej_Kristan,VOT_TPAMI,vot2021Kristan2021a}.
These sequences feature multiple objects and many occlusions, which are the scenarios we wish to address.
The NeighborTrack algorithm can be used with various S-networks, such as Ocean~\cite{ocean_zhang2020}, TransT~\cite{TransT} and OSTrack~\cite{ostrack}, to effectively overcome appearance changes and occlusions.
In addition, we use the mainstream medium and long-term datasets of the SOT field, including LaSOT~\cite{lasotDBLP:journals/corr/abs-1809-07845} and GOT-10K~\cite{got10kHuang2021}, to verify NeighborTrack with the state-of-the-art method, OSTrack~\cite{ostrack}.
VOT2020 and VOT2021 are mask datasets, while LaSOT, VOT2022bbox, and GOT-10K are bbox datasets.
We apply NeighborTrack to three previous models: a traditional attention-based network Ocean~\cite{ocean_zhang2020}, a transformer-based network TransT~\cite{TransT}, and a state-of-the-art network OSTrack~\cite{ostrack}.

\subsection{VOT datasets} 
\label{sec:VOT}

Several quantitative metrics are provided in the annual VOT challenge~\cite{VOT_TPAMI} for comparing the performance of SOT methods.
The metrics include (1) accuracy: the average IoU between the tracker output and the ground truth before tracking failure, (2) robustness: the percentage of successfully tracked sub-sequence frames, (3) Expected Average Overlap (EAO): the expected value of IoU between each tracking result and the ground truth.
For all metrics, higher is better. 
For accuracy, because we calculate the average IoU only when tracking is successful, we cannot determine whether the result is good or bad when tracking fails; the robustness metric, on the other hand, only reacts when tracking fails.
Compared to these metrics, the EAO metric is a principled combination of tracking accuracy and robustness, thus providing a more comprehensive picture of whether tracking has been successful or not.
Therefore, EAO has usually been used as the primary performance measure.

Table \ref{tabular:vot} shows the results of the proposed NeighborTrack.
NeighborTrack is used to augment three representative models, Ocean, TransT, and OSTrack.
Overall, NeighborTrack improves EAO by ${1.92\%}$ and robustness by ${2.11\%}$ on average.
The accuracy decreases slightly at ${-0.07\%}$ because this metric only considers the average IoU of successful tracking, while the frames that fail to track do not participate in its calculation.
In an extreme case, if the tracking fails in the second frame after the start, the accuracy calculation will reach the maximum of 1.0 since only the ground truth is used in the calculation process. However, this cannot be considered successful tracking.
With our method, tracking will be less likely to be interrupted, but the detected bounding box position could be inaccurate if the target is obscured.
The other two metrics, robustness and EAO, demonstrate that our method has the ability to help SOT models track the target more consistently and effectively.
In particular, for the primary metric EAO, NeighborTrack improves all three models on all datasets. 
The results show that our method is agnostic and effective to different types of SOT models.

\figref{votexamples} illustrates the tracking results for some examples. 
In the figure, the green box represents the ground truth; the magenta box represents the original results of the baseline, TransT~\cite{TransT}; and the red box represents the NeighborTrack-corrected results.
In the Bag example, since the garbage bag is white, as is the license plate and window, a tracker is likely to be confused.
After applying NeighborTrack, although there will still be a small amount of false tracking, tracking can be resumed very quickly in contrast to TransT, which cannot be resumed after losing track.
Crabs1 is a more challenging example since the crabs have similar appearances, which can lead to false tracking when they are close to each other.
By introducing a mechanism of utilizing neighbors, our proposed method effectively reduces thetracking failure caused by appearance similarity.
As a final example, we examine Shacking, an example with a singer shacking.
TransT loses tracks as a result of being attracted to other objects, such as the arm, other people's heads, and finally the drum.
Our method briefly mistracks when the face is completely obscured by the hat, but returns to the correct position as soon as the face appears again.

\iffalse
表1 modelname使用黑色字的是baseline model，加上我們後處理方法為青色字，以baseline為基礎，再加上我們的方法之後效果提升則標上綠色，在此表我們證明兩點(1)根據有/沒有使用transformer來分類SOT方法，我們的方法適用於沒有transformer的方法\cite{ocean_zhang2020}，使用CNN進行特徵擷取並且用transformer關聯特徵的方法\cite{TransT}，與完全使用transformer進行特徵擷取與關聯特徵的方法\cite{ostrack}，可以看到所有的方法在robustness與EAO都有提升.(2)我們的方法能夠與其他後處理方法\cite{alpharefine}\cite{DIMP}一起使用，VOT2020與VOT2021是mask dataset，我們用的S-network本身都不能產生mask，首先我們使用S-network直接加上\cite{alpharefine}產生mask得到baseline(如TransT\cite{TransT}+AR\cite{alpharefine})，接下來以此為基礎加上我們的方法(如TransT\cite{TransT}+AR\cite{alpharefine}+ours,此例在VOT2020上提升了4.1$\%$robustness與3.3$\%$EAO)，可以發現robustness與EAO都有明顯的效果提升，另外Ocean\cite{ocean_zhang2020}本身包含\cite{DIMP}方法，同時觀察VOT2020與VOT2021與VOT2022bbox可以發現不管有沒有與AR一起使用，以Ocean為基底的網路加上我們的方法都能夠提升EAO.
\fi

\begin{figure*}[t]
  \centering
  %\fbox{\rule[-.5cm]{4cm}{4cm} \rule[-.5cm]{4cm}{0cm}}
  \includegraphics[scale=0.47]{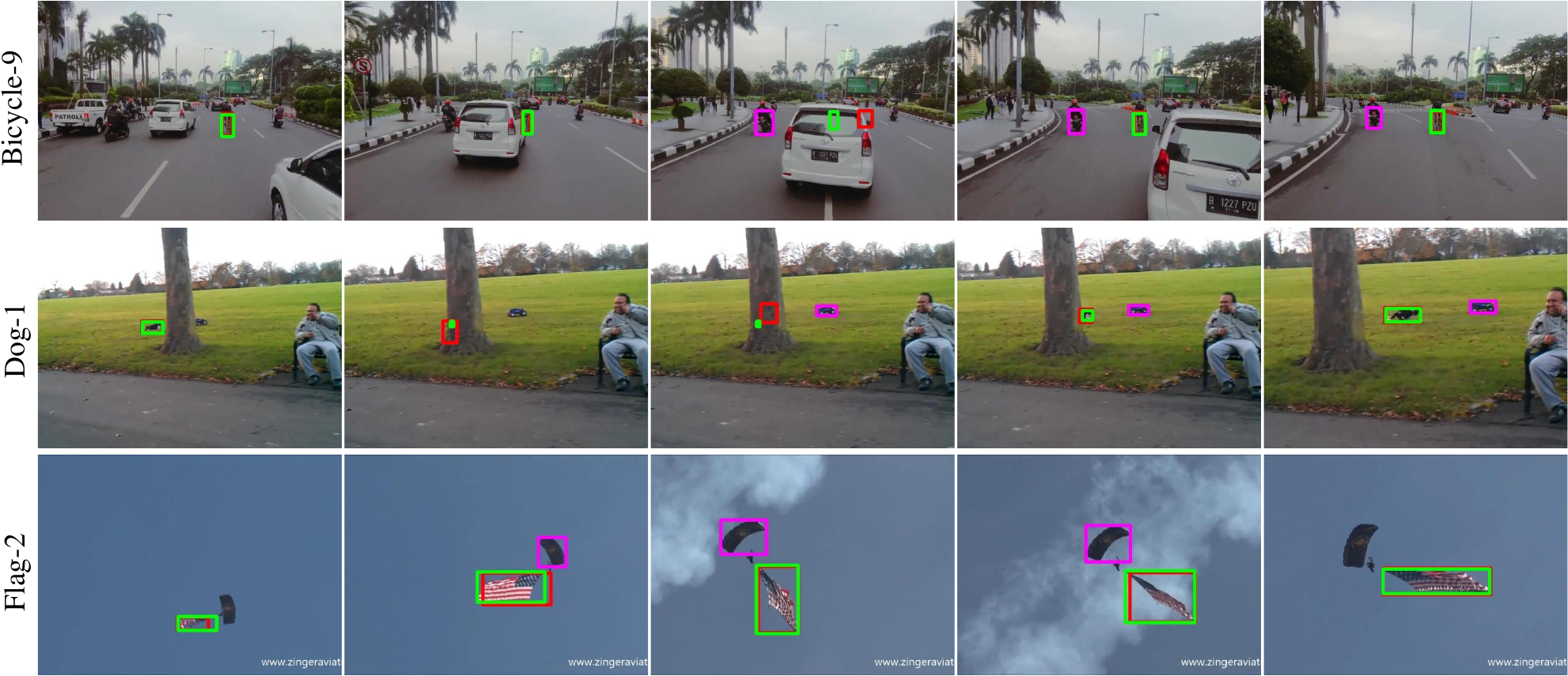}
  \caption{
  Examples of tracking results with and without NeighborTrack from LaSOT~\cite{lasotDBLP:journals/corr/abs-1809-07845}. 
  We apply NeighborTrack to correct the tracking results of OSTrack~\cite{ostrack}. 
  Ground truth is indicated by the green boxes.
  OSTrack's results are represented by magenta boxes, while NeighborTrack's results are represented by red boxes.
  OSTrack fails to track Bicycle-9 and Dog-1 due to the occlusion of the target, whereas it fails to track Flag-2 due to the target's deformation.
  NeighborTrack is able to correct the majority of tracking errors.
  }
  \iffalse
  \caption{利用NrighborTrack修正結果:baseline model為OSTrack\cite{ostrack}，綠色bbox為GT，紫色bbox為原始結果，紅色bbox為NrighborTrack修正後的結果。Bicycle-9與Dog-1是物體被遮蔽造成追蹤器選擇相似但錯誤的物體作為追蹤結果，Flag-2則是物體的型變造成外觀變化導致追蹤錯誤，經過我們的方法後都能追蹤正確的物體。資料庫為\cite{lasotDBLP:journals/corr/abs-1809-07845}}
  \fi
  \label{fig:lasotexamples}
\end{figure*}

\begin{table}
\caption{Comparisons of NeighborTrack (applied to OSTrack384) and other leading methods on the LaSOT~\cite{lasotDBLP:journals/corr/abs-1809-07845} and GOT-10K~\cite{got10kHuang2021} datasets.}
\centering
\scalebox{0.8}{
\begin{tabular}{cccc}

\toprule  %添加表格头部粗线
&&LaSOT&\\

model name& AUC$\uparrow$&	 Norm-Precision$\uparrow$& Precision$\uparrow$	\\

\midrule  %添加表格中横线
OSTrack384\cite{ostrack}  +ours 			&\color{red} 0.722& \color{red}0.818& \color{red}0.780\\
OSTrack384\cite{ostrack} 			& \color{green}0.711& \color{green}0.811 & \color{green}0.776\\
SwinV2-L 1K-MIM\cite{swintrackv2https://doi.org/10.48550/arxiv.2205.13543}        & \color{blue}0.707& --- & ---\\

SwinTrack-B-384\cite{SwinTracklin2021swintrack}        & 0.702& 0.784 & 0.753\\
MixFormer-L\cite{mixformer} 			& 0.701& \color{blue}0.799 &\color{blue} 0.763\\
AiATrack\cite{AiATrackgao2022aiatrack}        & 0.690& 0.794 & 0.738\\
Unicorn\cite{Unicornnetyan2022towards} 			& 0.685& 0.766 & 0.741\\

KeepTrack\cite{keepTrackmayer2021learning}        & 0.671& 0.772 & 0.702\\
DMTrack\cite{sotreidZhang_2021_CVPR} 			& 0.574& --- & 0.580\\

%OSTrack384\cite{ostrack}  +ours n18 			&0.721 &0.816 & 0.779\\
%TransTN4\cite{TransT} 				& 0.649& 0.738& 0.690\\
%TransTN2\cite{TransT}   			& 0.639& 0.732& 0.679\\
%TransTN2\cite{TransT}  +ours 			& 0.627& 0.717& 0.663\\
%TransTN2\cite{TransT}  +ours n18 			& 0.634& 0.726& 0.672\\

%Ocean\cite{ocean_zhang2020}	& 0.527& 0.610& 0.531\\
%Ocean\cite{ocean_zhang2020}  +ours 				& \color{green}0.529&\color{green} 0.612& \color{green} 0.530\\

\midrule  %添加表格中横线
&& GOT-10K&\\
model name& AO$\uparrow$&	 $SR_{0.5}\uparrow$& $SR_{0.75}\uparrow$	\\

\midrule  %添加表格中横线
OSTrack384\cite{ostrack}  +ours 			&\color{red} 0.757& \color{green}0.8572& \color{red}0.733\\
MixFormer-L\cite{mixformer} 			& \color{green}0.756& \color{red}0.8573 & \color{green}0.728\\
OSTrack384\cite{ostrack} 			& \color{blue}0.737& \color{blue}0.832 & \color{blue}0.708\\
SwinV2-L 1K-MIM\cite{swintrackv2https://doi.org/10.48550/arxiv.2205.13543}        & 0.729& --- & ---\\

SwinTrack-B-384 \cite{SwinTracklin2021swintrack}        & 0.724& 0.805 & 0.678\\
%MixFormer-1K\cite{mixformer} 			& 0.712& 0.799 & 0.658\\
%MixFormer-22K\cite{mixformer} 			& 0.707& 0.800 & \color{blue}0.678\\
AiATrack\cite{AiATrackgao2022aiatrack}        & 0.696& 0.800 & 0.632\\

STARK\cite{STARKyan2021learning}        & 0.688& 0.781 & 0.641\\
%SwinTrack-B\cite{SwinTracklin2021swintrack}        & 0.694& 0.780 & 0.643\\
\midrule  %添加表格中横线

%&& Total AVG&\\
%\midrule  %添加表格中横线
%+ours 						& \color{green}${(+0.65\%)}$& \color{green} ${(+0.45\%)}$& \color{green} ${(+0.4\%)}$\\

\bottomrule %添加表格底部粗线

\end{tabular}}
\label{tabular:long_lasot}
\end{table}

\subsection{LaSOT and GOT-10K datasets} 
\label{sec:LaSOT}

For the experiments of medium-term and long-term tracking, we used two datasets, LaSOT~\cite{lasotDBLP:journals/corr/abs-1809-07845} and GOT-10K~\cite{got10kHuang2021}.
We use the metrics suggested by each dataset. 
Table \ref{tabular:long_lasot} summarizes the results.
The top three results are colored red, green, and blue, respectively.
For LaSOT, OSTrack384~\cite{ostrack} leads all other methods. 
When NeighborTrack is applied to OSTrack384, its performance is further improved, and OSTrack384+NeighborTrack outperforms all other methods in all metrics.
The performance of OSTrack384 on GOT-10K is substantially inferior to MixFormer-L~\cite{mixformer}, the best method excluding ours. 
OSTrack384+NeighborTrack, however, outperforms MixFormer-L, because NeighborTrack boosts OSTrack384 significantly.
Accordingly, NeighborTrack achieves state-of-the-art results for both the LaSOT and GOT-10K datasets.

\figref{lasotexamples} provides examples of tracking results for LaSOT.
In Bicycle-9, a bicycle is tracked as it travels in an array of vehicles.
When a white car obscures the bicycle, the tracker tracks a similar-looking black motorcycle, since it loses its target. 
NeighborTrack recognizes that the black motorcycle is not the intended target, so it is set up to re-track the target at a later time.
Our method has succeeded in this example by introducing the Kalman filter, which provides candidates independent of appearance features, thus providing a greater range of options to the tracking system.
Although the Kalman filter is not extremely accurate, it can serve as a guide to prevent the tracker from becoming outrageous.
Dog-1 involves tracking a dog.
When the dog is obscured by a tree, the tracker mistakenly selects the remote control car as the target.
NeighborTrack detected a wrong target and switched to a near-blocker object framed by a Kalman filter-generated box.
Flag-2 shows a typical example of appearance change due to deformation.
In this case, the target to be tracked is a towed flag whose appearance varies greatly over time because of the soft nature of the flag and its variable pattern.
Traditional methods can cause confusion between the flag and the parachute adjacent to it.
Using NeighborTrack, as long as one of the two sides is relatively stable, neighbor information should be useful in reducing tracking failures caused by such deformation.

%% file: 5conclusion.tex
\section{Conclusion}
\label{sec:conclusion}

We propose NeighborTrack, a post-processing scheme that is agnostic and can be applied to state-of-the-art single object tracking methods provided that confidence scores are available.
Through the use of neighbor information, NeighborTrack effectively mitigates the appearance change problem caused by occlusion or deformation.
Our approach can be applied to both non-transformer-based methods~\cite{ocean_zhang2020} and transformer-based methods~\cite{TransT,ostrack}.
Extensive experiments demonstrate that the proposed method is capable of improving the tracking performance of various SOT methods.
There is no need to collect any additional training data for the proposed method.
Furthermore, there is no need to retrain the SOT network.
In summary, NeighborTrack complements most current state-of-the-art SOT algorithms and improves their accuracy.

%% file: 6additional_result.tex
\section{Additional Result}
\label{sec:additional_result}

\iffalse
本章節顯示未包含在原始會議論文中的其他結果，我們結合OSTrack測試於UAV123\cite{uav123_2016},TrackingNet\cite{trackingnet_2018},OTB100\cite{OTB2015}，UAV123與OTB100的測試結果如表\ref{tabular:additional result}所示，TrackingNet的測試結果如表\ref{tabular:additional result trackingnet}所示,可以發現隨著$\tau$增加，AUC就跟著增加。應注意的是，由於UAV123的影像多屬於長距離跟蹤，若$\tau$過低會造成效果下降。

\fi

This section presents additional results that were not included in the original conference paper.
We tested OSTrack on UAV123 \cite{uav123_2016},
TrackingNet \cite{trackingnet_2018},
and OTB100 \cite{OTB2015}.
The test results for UAV123 and OTB100 are shown in Table \ref{tabular:additional result},
and the results for TrackingNet are presented in Table \ref{tabular:additional result trackingnet}.
It is observed that as the value of $\tau$ increases,
the Area Under Curve (AUC) also increases.
It should be noted that,
since most of the UAV123 videos involve long-distance tracking,
a low value of $\tau$ can lead to decreased performance.

\begin{table}
\caption{Comparisons of NeighborTrack (applied to OSTrack384) and other leading methods on the UAV123\cite{uav123_2016} and OTB100\cite{OTB2015} datasets.}
\centering
\scalebox{0.8}{
\begin{tabular}{ccccccc}

\toprule  %添加表格头部粗线
&&&UAV123&&&\\

model name& AUC$\uparrow$&	 OP50$\uparrow$&	 OP75$\uparrow$&	 Precision$\uparrow$& Norm Precision$\uparrow$& FPS$\uparrow$	\\

\midrule  %添加表格中横线
OSTrack384\cite{ostrack}  +ours(\(\tau = 27\)) &\color{red} 0.7256& \color{red}0.8775& \color{red}0.6815&\color{red} 0.9337&\color{red} 0.8851&\color{blue} 1.31\\

OSTrack384\cite{ostrack}  +ours(\(\tau = 9\)) &\color{blue} 0.7152& \color{blue}0.8641& \color{blue}0.6747&\color{blue} 0.9186&\color{blue} 0.8727&\color{green} 2.11\\

OSTrack384\cite{ostrack} 			&\color{green} 0.7217& \color{green}0.8724& \color{green}0.6809&\color{green} 0.9259&\color{green} 0.8806&\color{red} 3.83\\

\midrule  %添加表格中横线
		
&&& OTB100&&&\\
model name& AUC$\uparrow$&	 OP50$\uparrow$&	 OP75$\uparrow$&	 Precision$\uparrow$& Norm Precision$\uparrow$& FPS$\uparrow$	\\

\midrule  %添加表格中横线
OSTrack384\cite{ostrack}  +ours(\(\tau = 27\))&\color{red} 0.6974& \color{red}0.8588& \color{red}0.5649&\color{red} 0.9042&\color{red} 0.8487&\color{blue} 1.23\\

OSTrack384\cite{ostrack}  +ours(\(\tau = 9\)) &\color{green} 0.6954& \color{green}0.8552& \color{green}0.5640&\color{green} 0.9021&\color{green} 0.8468&\color{green} 1.98\\

OSTrack384\cite{ostrack} 			&\color{blue} 0.6927& \color{blue}0.8542& \color{blue}0.5639&\color{blue} 0.8962&\color{blue} 0.8438&\color{red} 3.91\\

\bottomrule %添加表格底部粗线

\end{tabular}}
\label{tabular:additional result}
\end{table}

\begin{table}
\caption{Comparisons of NeighborTrack (applied to OSTrack384) and other leading methods on the TrackingNet\cite{trackingnet_2018} datasets.}
\centering
\scalebox{0.8}{
\begin{tabular}{cccc}

\toprule  %添加表格头部粗线
&& TrackingNet&\\
model name& Success	$\uparrow$&	 Precision$\uparrow$& Normalized Precision $\uparrow$	\\

\midrule  %添加表格中横线
OSTrack384\cite{ostrack}  +ours(\(\tau = 18\)) 			&\color{red}0.8379 & \color{red}0.8324& \color{red}0.8830\\
OSTrack384\cite{ostrack}  +ours(\(\tau = 9\)) 			&\color{green}0.8373 & \color{green}0.8316& \color{green}0.8823\\
OSTrack384\cite{ostrack} 			& \color{blue}0.8358& \color{blue}0.8294 & \color{blue}0.8805\\

\bottomrule %添加表格底部粗线

\end{tabular}}
\label{tabular:additional result trackingnet}
\end{table}